\documentclass[letterpaper]{article} 
\usepackage{aaai2026}  
\usepackage{times}  
\usepackage{helvet}  
\usepackage{courier}  
\usepackage[hyphens]{url}  
\usepackage{graphicx} 
\usepackage{amsmath}
\usepackage{xcolor}
\usepackage{multicol}
\usepackage{multirow}
\usepackage{amsmath}
\usepackage{amssymb}

\usepackage{natbib}  
\usepackage{caption} 
\usepackage{algorithm}
\usepackage{algorithmic}
\usepackage{booktabs}

\usepackage{makecell}

\usepackage{newfloat}
\usepackage{listings}
\DeclareCaptionStyle{ruled}{labelfont=normalfont,labelsep=colon,strut=off} 
\floatstyle{ruled}
\newfloat{listing}{tb}{lst}{}
\floatname{listing}{Listing}
\title{BCWildfire: A Long-term Multi-factor Dataset and Deep Learning Benchmark for Boreal Wildfire Risk Prediction}
\author{
 Zhengsen Xu\textsuperscript{\rm 1}, 
 Sibo Cheng\textsuperscript{\rm 2}, 
 Lanying Wang\textsuperscript{\rm 3}, 
 Hongjie He\textsuperscript{\rm 3, 4}, 
 Wentao Sun\textsuperscript{\rm 5}, 
 Jonathan Li\textsuperscript{\rm 3, 4, 5}, 
 Lincoln Linlin Xu\textsuperscript{\rm 1}\thanks{Corresponding author.}
}
\affiliations{
    \textsuperscript{\rm 1}Schulich School of Engineering, University of Calgary, Canada\\
    \textsuperscript{\rm 2}CEREA, ENPC, EDF R\&D, Institut Polytechnique de Paris, France\\
    \textsuperscript{\rm 3}Department of Environmental Manage, University of Waterloo, Canada\\
    \textsuperscript{\rm 4}Hinton STAI Institute, East China Normal University, China\\
    \textsuperscript{\rm 5}Department of Systems Design Engineering, University of Waterloo, Canada\\

    \{zhengsen.xu, lincoln.xu\}@ucalgary.ca, sibo.cheng@enpc.fr, \{hongjie.he, lanying.wang, wentaolsun, junli\}@uwaterloo.ca
}

\begin{document}

\maketitle

\begin{abstract}
Wildfire risk prediction remains a critical yet challenging task due to the complex interactions among fuel conditions, meteorology, topography, and human activity. Despite growing interest in data-driven approaches, publicly available benchmark datasets that support long-term temporal modeling, large-scale spatial coverage, and multimodal drivers remain scarce. To address this gap, we present a 25-year, daily-resolution wildfire dataset covering 240 million hectares across British Columbia and surrounding regions. The dataset includes 38 covariates, encompassing active fire detections, weather variables, fuel conditions, terrain features, and anthropogenic factors. Using this benchmark, we evaluate a diverse set of time-series forecasting models, including CNN-based, linear-based, Transformer-based, and Mamba-based architectures. We also investigate effectiveness of position embedding and the relative importance of different fire-driving factors.

\end{abstract}

\begin{links}
\link{Code and Datasets}{https://github.com/SynUW/BCWildfire}
\end{links}



\section{Introduction}

Wildfires have experienced unprecedented escalation in frequency, scale, and intensity globally, with climate change exacerbating fire-prone conditions and creating an urgent need for accurate wildfire risk prediction models~\cite{ager2019wildfire}. Boreal regions, characterized by substantial carbon storage and high wildfire susceptibility, face particularly acute risks \cite{stephens2014temperate,jain2024drivers}. For instance, British Columbia's wildfire patterns threaten both regional ecosystems and global climate stability, necessitating robust predictive frameworks for effective fire management strategies \cite{parisien2023abrupt,daniels20242023}.

Data-driven approaches, especially deep learning architectures, have shown great promise in wildfire prediction due to their capacity to model complex nonlinear interactions among heterogeneous environmental variables \cite{XU2025632}. However, the deployment of such models is often hindered by the lack of comprehensive and high-quality training datasets. Despite extensive research on wildfires, publicly available standardized datasets remain limited in terms of scope and accessibility \cite{gerard2023wildfirespreadts}.

\begin{figure}[t]
    \centering
    \includegraphics[width=1.0\linewidth]{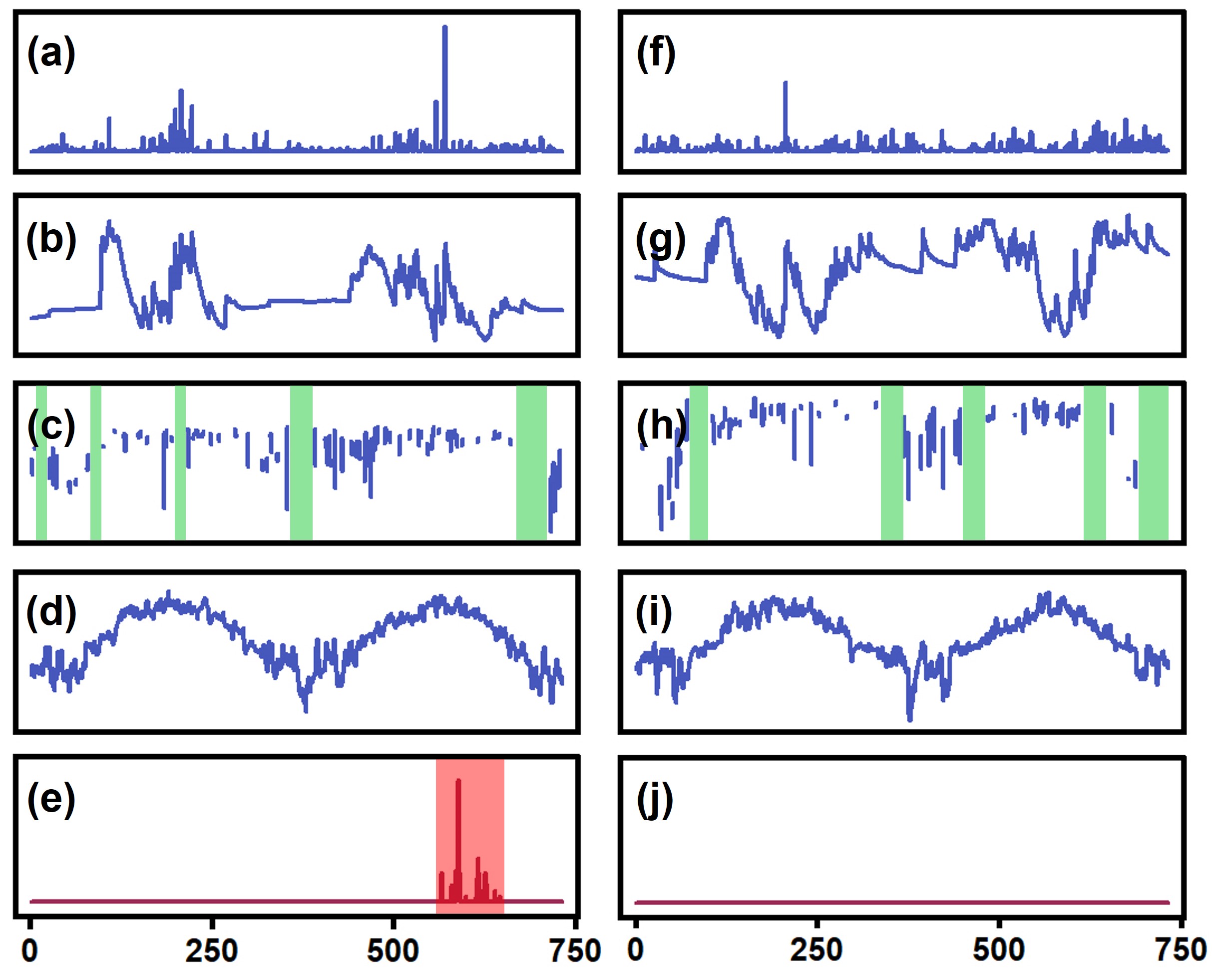}
    \caption{Time series of key variables (2023-2024) showing wildfire prediction challenges. Figure (a)-(e): precipitation, soil moisture L1, MODIS Band 20, 2m temperature (exogenous), and burned area (endogenous) at an ignition point. Figure (f)-(j): same variables in non-ignition areas. Red: wildfire periods; Green: missing data. The figure demonstrates spatiotemporal decoupling between exogenous/endogenous variables and remote sensing data gaps.}
    \label{fig:challendge_illustration}
\end{figure}

Current wildfire prediction systems face fundamental challenges across multiple dimensions. A key limitation lies in the lack of standardized, publicly accessible datasets with broad geographic coverage, particularly in carbon-rich boreal ecosystems. Most existing research focuses on localized fire spread modeling (see Table~\ref{tab:oadataset}), typically using input sequences constrained to short temporal windows such as 1 or 7 days \cite{eddin2023location,deng2025daily,zhou2025comparative}. This narrow temporal scope fails to account for large-scale, long-term wildfire risk driven by fuel accumulation, prolonged drought, and anthropogenic activity over seasonal to annual timescales. In addition, the integration of diverse and heterogeneous fire-driving factors remains limited, which restricts models' ability to capture the complex and multi-scale mechanisms that govern wildfire ignition and spread. 

\begin{table}[t]
\centering
\begin{tabular}{lcc}
\hline
Dataset & \makecell{Temporal Span\\(Lookback/Forecast)} & Resolution \\
\hline
FireCube & \makecell{Daily/Daily} & Daily \\
WildfireDB & \makecell{Daily/Daily} & Daily \\
WildfireSpreadTS & \makecell{14--94 d/4--94 d} & Daily \\
\makecell[l]{Next Day\\Wildfire Spread} & \makecell{Daily/Daily} & Daily \\
\makecell[l]{SeasFire Cube} & \makecell{N/A/N/A} & 8 d \\
\hline
BCWildfire (ours) & \makecell{$\leq$25 y/$\leq$25 y} & Daily \\
\hline
\end{tabular}
\caption{Comparison between current next day or event-based  wildfire prediction datasets and BCWildfire.}
\label{tab:oadataset}
\end{table}


By contrast, methodological challenges arise from the inherent complexity of wildfire systems, where risk is driven by long-term cumulative interactions among diverse geospatial covariates, including meteorological conditions (e.g., temperature, humidity, precipitation), vegetation characteristics (e.g., fuel load, moisture content), human activity, and topography \cite{XU2025632}. The interdisciplinary nature of this domain has led to a knowledge gap: machine learning practitioners often lack the geospatial expertise required to effectively apply advanced algorithms to wildfire prediction, while wildfire experts continue to rely on traditional models such as Random Forest and XGBoost \cite{di2025global}. As a result, data-driven wildfire research has yet to fully leverage the recent advances in deep learning.

To address these fundamental limitations, we present BCWildfire, a comprehensive spatiotemporal dataset that covers British Columbia, Canada, and adjacent regions, spanning approximately 240 million hectares over the period 2000 to 2024 and incorporating 38 multimodal wildfire driving factors. We further conduct a systematic benchmark of deep learning architectures, including CNN-based, linear-based, Transformer-based, and Mamba-based models, for the next day wildfire risk prediction to evaluate their effectiveness on this challenging task.

The primary contributions of this work are:
\begin{itemize}
\item We introduce BCWildfire, a large-scale spatiotemporal dataset for wildfire risk prediction, consisting of 2.4 million samples across 240 million hectares of boreal landscapes over a 25-year period. The dataset integrates 38 multimodal wildfire-driving factors, addressing major limitations in existing resources related to geographic coverage, feature completeness, and long-term temporal representation.

\item In contrast to existing datasets that primarily focus on wildfire spread and behavior modeling based on spatial information, BCWildfire is specifically designed to support time series forecasting. This design enables the modeling of cumulative interactions among wildfire drivers over time, thereby facilitating the development of models with enhanced temporal reasoning capabilities and the potential to establish new benchmarks for wildfire prediction.

\item We conduct a systematic benchmarking of state-of-the-art deep learning architectures across 4 major paradigms including CNN-based, linear-based, Transformer-based, and Mamba-based models for spatiotemporal wildfire prediction. Our evaluation highlights the strength and limitations of each paradigm in handling complex variable dependencies, non-periodic patterns, and high stochastic inherent in wildfire dynamics.
\end{itemize}

\section{Related Work}
\subsection{Datasets}

Several open-source wildfire datasets have been released in recently, including WildfireDB~\citep{singla2020wildfiredb}, Next Day Wildfire Spread~\citep{huot2022next}, FireCube~\citep{prapas_2022_6475592, kondylatos2023mesogeos}, WildfireSpreadTS~\citep{gerard2023wildfirespreadts}, and SeasFire Cube~\citep{alonso_2023_8055879} (see Table~\ref{tab:oadataset}). These datasets compile environmental covariates related to wildfire events across the United States and the Mediterranean, incorporating meteorological conditions, topography, vegetation (fuel) status, human activity, and fire behavior. Most of these datasets are designed to support models that use recent covariate states and fire activity as inputs to predict wildfire spread at the next time step. Consequently, as summarized in Table~\ref{tab:oadataset}, most studies generally adopt a daily temporal resolution with short look back periods and limited prediction horizons. Moreover, they mainly emphasize small scale wildfire spread and behavior prediction after ignition, rather than large scale wildfire risk prediction.

Overall, the number and spatiotemporal coverage of publicly available wildfire datasets remain limited. Existing datasets tend to focus on fire spread modeling for individual events within short time windows, primarily in the U.S. and Mediterranean regions. This limited scope constrains the development of wildfire ignition risk prediction models, which rely on long-term accumulative drivers and are crucial for proactive wildfire management \cite{XU2025632}. Moreover, due to the high sensitivity of wildfire behavior to local environmental and climatic conditions, models trained on region-specific or structurally constrained datasets often fail to generalize to boreal forests \cite{gerard2023wildfirespreadts}. 

From a time series forecasting perspective, although there exists a wide range of datasets in domains such as electricity trading, energy consumption, traffic flow, and weather forecasting, there remains a notable scarcity of datasets focused on natural disaster prediction, particularly those involving complex and stochastic hazards like wildfires \cite{chen2001freeway,lai2018modeling,wu2021autoformer,zhou2021informer}. This scarcity limits the ability of machine learning researchers to apply and evaluate advanced time series forecasting algorithms in real-world disaster risk scenarios.

\begin{figure*}
    \centering
    \includegraphics[width=1\linewidth]{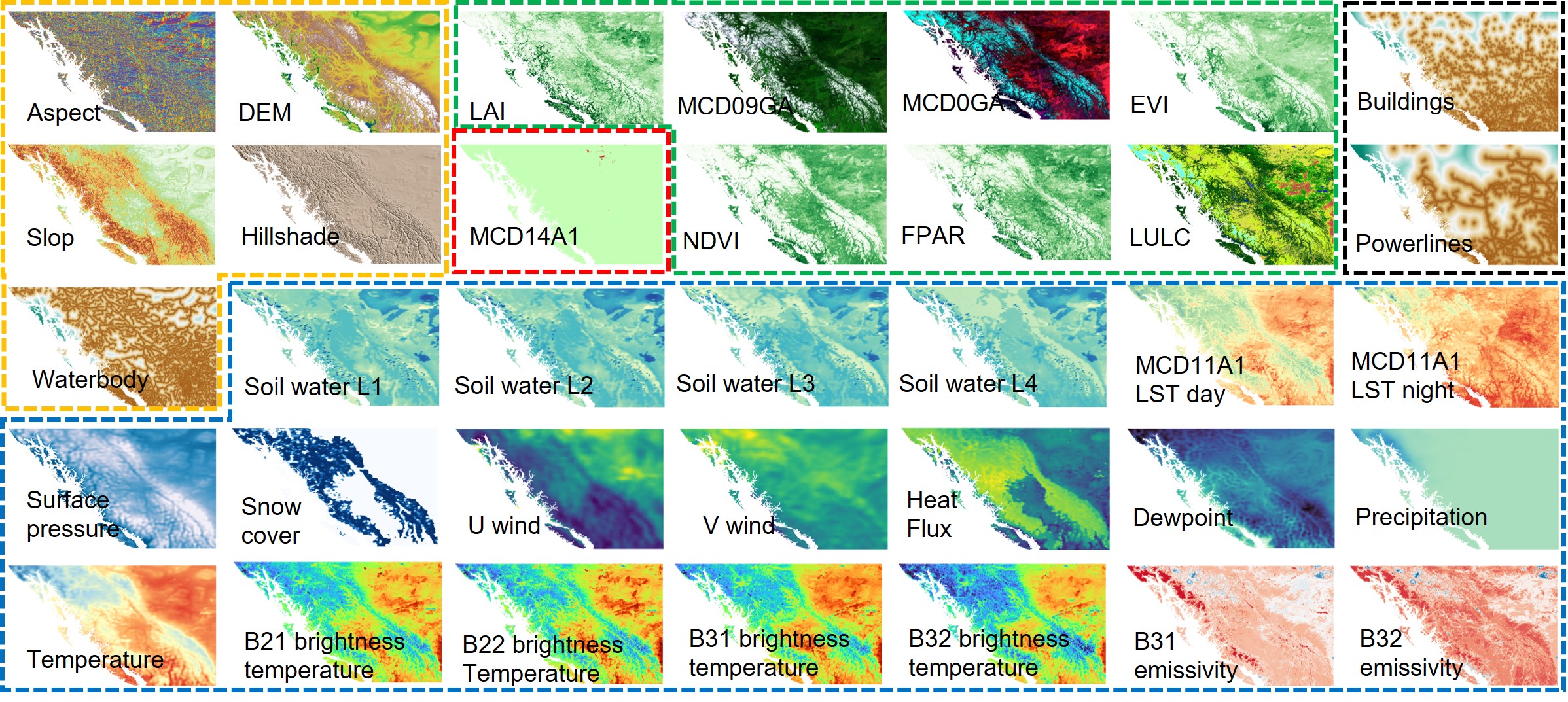}
    \caption{Partial overview of the dataset, illustrating fuel (green), fire detection (red), topography (yellow), human activities (black), and meteorological factors (blue).}
    \label{fig:datasetoverview}
\end{figure*}

\subsection{Methods}

Current wildfire risk prediction research predominantly treats the problem as a spatiotemporal forecasting task, utilizing wildfire burning states and covariate conditions from previous days as inputs to predict wildfire coverage extent and behavior at next day \cite{10149031,bhowmik2023multi,ali2024advancing}. Consequently, algorithmic approaches frequently employ various CNN and Vision Transformer-based variants, such as UNet, UNETR, and T4-Fire, with primary emphasis on spatial information consideration, while neglecting the cumulative effects of long-term driving factors \cite{ronneberger2015u,hatamizadeh2022unetr,zhao2024tssatfiremultitasksatelliteimage}. On the other hand, wildfire prediction algorithms based on long-term time series information remain predominantly grounded in traditional machine learning methodologies, including XGBoost, Random Forest, simple feedforward neural networks, and Long Short-Term Memory (LSTM) \cite{kondylatos2022wildfire,di2025global}.

Deep learning has greatly advanced time series forecasting, enabling models to infer future dynamics from historical observations. Multiple model architectures have emerged based on CNN, Linear, Transformer, and Mamba frameworks \cite{wang2025mamba}, including CNN-based TimesNet \cite{wu2022timesnet} and SCINet \cite{liu2022scinet}, Linear-based TiDE \cite{das2024longtermforecastingtidetimeseries}, and CrossLinear \cite{zhou2025crosslinear}, as well as Transformer-based models such as TimesNet \cite{wu2022timesnet}, Autoformer \cite{wu2021autoformer}, FEDformer \cite{zhou2022fedformer}, Crossformer \cite{zhang2023crossformer}, PatchTST \cite{nie2023a}, and iTransformer \cite{liu2024itransformer}. These models have demonstrated significant advantages across diverse prediction tasks in transportation, meteorology, and trading domains \cite{wang2025mamba,zhou2025crosslinear}. However, exploration in natural disaster applications, particularly in wildfire risk prediction, remains notably limited.


\section{BCWildfire Benchmark Dataset}

\subsection{Overview}

Overall, we construct a long-term time series forecasting dataset with consistent spatiotemporal resolution, covering British Columbia, Canada, and surrounding regions from 2000 to 2024. Each sample includes 38 daily covariates grouped into five categories: vegetation (fuel) conditions, active fire detections, meteorological factors, human activity, and topographic features, as shown in Figure~\ref{fig:datasetoverview}. The dataset primarily sources data from MODIS satellite remote sensing products, ERA5-Land reanalysis data \cite{munoz2021era5}, OpenStreetMap \cite{openstreetmap2021}, and the ASTER digital elevation model (DEM) \cite{jarvis2004practical}. As illustrated in Figure~\ref{fig:datasetoverview}, each day's observation is organized into a structured data cube with all features aligned. Given the varying native resolutions of the data sources, we resample all inputs to a uniform spatial resolution of 1 km and a daily temporal resolution to ensure consistency across both dimensions.

\subsection{Driving Factors}

\subsubsection{Fuel Conditions}

Vegetation serves as the primary fuel for wildfires; therefore, capturing its dynamic changes is essential for predicting wildfire occurrence and spread \cite{smith2023there}. Given the rapid dynamics of wildfire behavior, stable products that provide long-term time series with high temporal resolution are extremely scarce. Accordingly, we selected MODIS products to characterize fuel conditions. First, we used the stable 500 m, 4-day Leaf Area Index (LAI) composite product (MCD15A3H), which provides LAI and the Fraction of Photosynthetically Active Radiation (FPAR) \cite{myneni2015mcd15a3h}. These variables serve as proxies for fuel load, structure, moisture, and vegetation activity. In addition, to better capture the rapid temporal dynamics of vegetation, we employed the daily 500 m MOD/MYD09GA reflectance data, specifically Bands 1, 2, 3, and 7, together with the derived Normalized Difference Vegetation Index (NDVI) and Enhanced Vegetation Index (EVI), to jointly represent fuel conditions \cite{vermote2015mod09ga}. The combination of these bands and products not only provides robust proxies for fuel status but also effectively distinguishes healthy vegetation from burned areas while maintaining sensitivity to clouds and smoke \citep{gerard2023wildfirespreadts}.

\subsubsection{Meteorological Factors}

Weather and climate factors are primary drivers of wildfire activity, influencing fire occurrence, burned area, and fire behavior, thus occupying a central position in various wildfire prediction models \cite{XU2025632}. High temperature, low relative humidity, insufficient precipitation, and strong winds are key short-term meteorological conditions affecting fire occurrence and spread \cite{kondylatos2022wildfire}. This study employs daily composite ERA5-Land products (with approximately 11 km spatial resolution) to extract meteorological driving factors \cite{munoz2021era5}. In variable selection, beyond key factors that directly influence fuel moisture, wind transport, and fire propagation, such as 2 m air temperature, 10 m eastward and northward wind components, atmospheric pressure, total precipitation, surface latent heat flux, and 2 m dewpoint temperature, we additionally include snow cover as a seasonal variable with a significant impact in northern regions. Snow cover not only modulates surface temperature and humidity but also significantly delays the exposure and drying process of combustible materials during winter and spring seasons, thereby affecting wildfire occurrence probability and initial spread rate. Furthermore, to more comprehensively reflect the cumulative effects of environmental humidity over longer time scales, this study also incorporates soil moisture data from the 0–289 cm depth range.

It is important to note that the relatively coarse spatial resolution of ERA5-Land leads to spatially smoothed temperature fields, which are insufficient to capture fine-scale thermal anomalies associated with wildfire ignition and spread. To overcome this limitation, we additionally incorporate MODIS thermal products, including MOD/MYD11A1 and MOD/MYD09CMG, with spatial resolutions of approximately 1 km and 5.6 km, respectively \cite{wan2021mod11a1_v061,vermote2015mod09cmg}. The MOD/MYD11A1 product provides day- and night-time land surface temperature and emissivity derived from Bands 31 and 32, whereas the MOD/MYD09CMG product contains brightness temperature from Bands 20, 21, 31, and 32. By combining these MODIS products with ERA5-Land temperature data, we aim to enhance the spatial detail and responsiveness to local thermal anomalies.

\subsubsection{Topographical Factors}

Topographical factors also play an important role in the ignition and spread of wildfires \cite{coen2013wrf}. Steep slopes help fire to rise quickly uphill, increasing the rate of spread; aspect affects sunlight exposure and dryness, thereby influencing the flammability of fuels; and hillshade reflects, to some extent, the shading of terrain on solar radiation, indirectly affecting local microclimate and burning conditions \cite{deng2025daily}. Therefore, we selected 30-meter resolution ASTER DEM data and calculated slope, aspect, and hillshade based on it \cite{jarvis2004practical}. Additionally, water bodies often serve as natural barriers that can significantly reduce the probability of fire spread, so we also calculated the distance to the nearest water body using OpenStreetMap's water distribution map, further enhancing the model's ability to represent fire behavior \cite{openstreetmap2021}.

\subsubsection{Human Activity Factors}

Human activities represent a major source of wildfire ignition \cite{bowring2024road}. To capture their potential influence, we incorporated the yearly 500 m resolution MODIS Land Use product (MCD12Q1) to account for the impact of different land use types on wildfire occurrence, as substantial contrasts exist between agricultural and forested areas in terms of fuel characteristics and human intervention \cite{friedl2019mcd12q1}. In addition, using settlement and power grid distribution data from OpenStreetMap, we computed the distance to the nearest infrastructure for each pixel. Since human activities tend to cluster around infrastructure, such proximity may increase the likelihood of ignition events \cite{openstreetmap2021}.

\subsubsection{Wildfire Data}

To construct temporally continuous and spatially consistent records of wildfire activity over the extended study period, we utilize daily MODIS Active Fire products MOD/MYD14A1, both with 1 km spatial resolution \cite{giglio2015mod14a1}. These products are based on thermal anomaly detection in mid-infrared bands (primarily Bands 20 and 21), providing global coverage with up to four observations per day. MODIS Active Fire products offer several advantages, including extensive historical coverage, high temporal frequency, and consistent detection algorithms, making them particularly suitable for large-scale and long-term fire trend analysis. We merge MOD and MYD detection data daily to generate composite active fire masks, retaining only pixels with high confidence levels to reduce false positives. When a pixel is detected as actively burning, its value reflects the most recent observation time for that day. This approach reliably identifies fire location and intensity, enabling precise delineation of wildfire ignition timing and spatial extent. The resulting time series provides a robust foundation for downstream tasks such as wildfire risk modeling and fire spread analysis.

\subsection{Dataset Preparation}

\begin{figure*}[htbp]
\centering
\includegraphics[width=\textwidth]{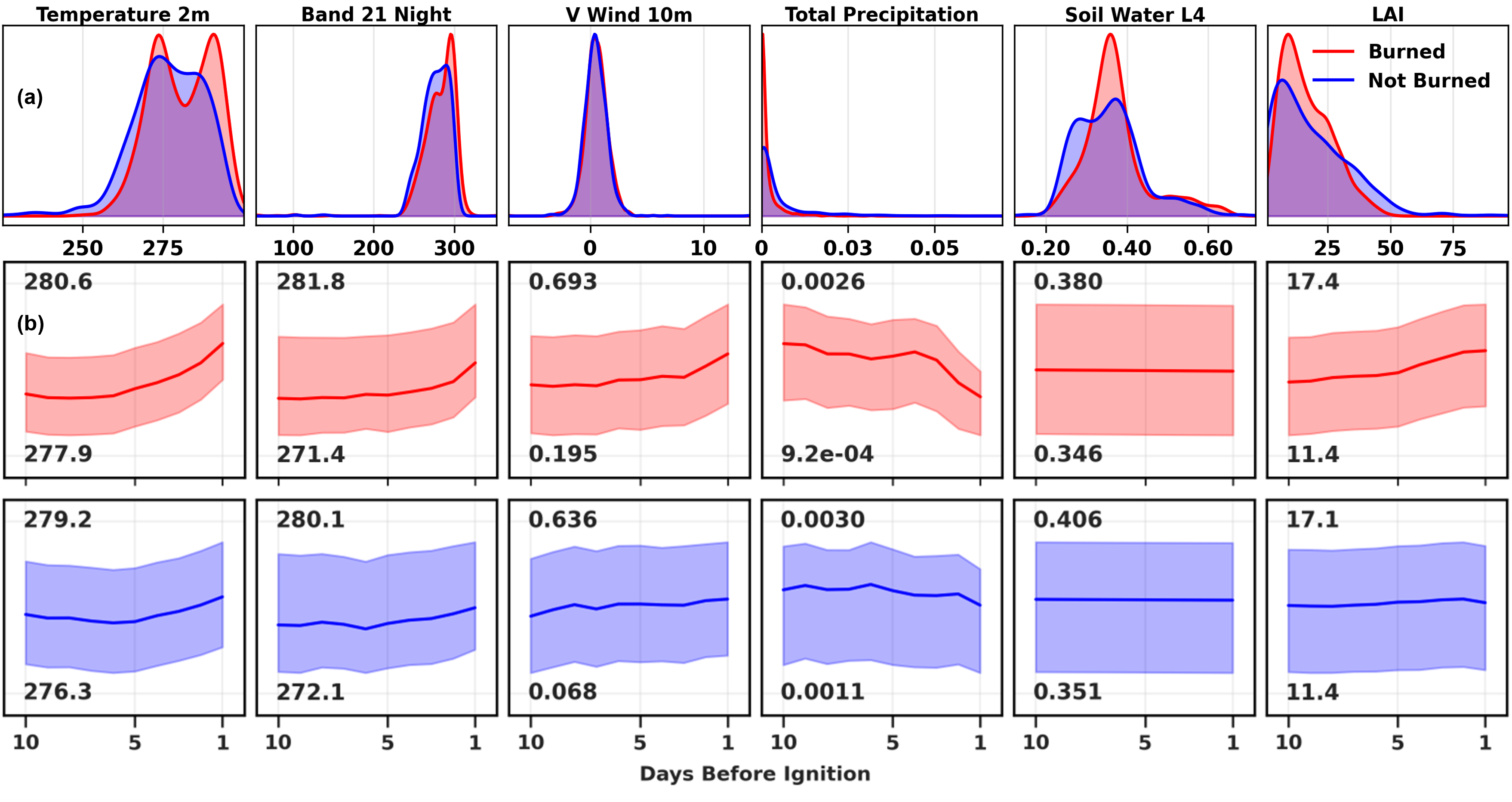}
\caption{Comparison of key environmental drivers between burned and unburned areas prior to wildfire ignition. (a) Kernel density distributions of six selected variables within 10 days before ignition. (b) Mean temporal trends of the same variables.}
\label{fig:distribution}
\end{figure*}

As our dataset integrates multimodal drivers from diverse sources, several challenges arise during its construction, including differences in geographic reference systems, spatial and temporal resolutions, and cloud contamination in optical remote sensing imagery. Moreover, since wildfires are extreme and relatively rare events, their spatiotemporal distribution is highly sparse compared to non-fire occurrences. Consequently, when wildfire risk prediction is formulated as a binary classification problem, it suffers from severe class imbalance between positive (fire) and negative (non-fire) samples.

The study area is frequently influenced by maritime air masses. Moist westerlies from the Pacific Ocean are forced to ascend over the Rocky Mountains, resulting in orographic precipitation and persistent cloud accumulation, which cause substantial cloud contamination and data gaps in MODIS optical imagery (see Figure~\ref{fig:challendge_illustration}). To mitigate this issue, we applied the quality control (QC) bands provided by the MODIS products to mask low-quality observations, followed by temporal filling using historical records. This approach produces cloud-free observations without introducing future data leakage.

Given the substantial discrepancies in coordinate reference systems and spatiotemporal resolutions among different drivers, all spatial inputs were standardized to the WGS84 coordinate system to ensure geographic consistency. We further resampled all inputs to a uniform 1~km grid, using nearest-neighbor interpolation for categorical data (e.g., land use and fire occurrence) and bilinear interpolation for continuous variables. To address temporal heterogeneity, we forward-filled non-daily data sources under the assumption of short-term temporal stability, thereby aligning all variables into a unified daily sequence.

Ultimately, we constructed a large-scale, preprocessed wildfire risk assessment dataset in GeoTIFF format, covering British Columbia and adjacent regions from 2000 to 2024. The dataset comprises 38 daily variables over a spatial domain of 2,782~$\times$~1,302~km$^2$ at 1~km resolution. All files share identical spatial extent and resolution across and within variables, providing a flexible structure that facilitates customized preprocessing and dataset assembly.

In our experiments, wildfire risk prediction is formulated as a binary time-series prediction problem, where the goal is to predict the probability of wildfire occurrence at time $t$ based on historical fire ignition and driver conditions from $t-1$ to $t-n$ at the pixel level. Because wildfires are rare events, negative samples (non-fire pixels) greatly outnumber positive samples (fire pixels). To alleviate this imbalance, we adopted an undersampling strategy~\cite{kondylatos2025uncertaintyawaredeeplearningwildfire}. In this benchmark, rather than randomly selecting negative samples, we excluded those located within a 60~km and 3-day spatiotemporal buffer around positive samples to avoid sampling from high-risk regions. Following the experimental setup of \citeauthor{kondylatos2022wildfire}, we ensured that for each land cover type and across all years, the number of negative samples remained a fixed multiple of the positive samples (two times in the training and validation sets, and equal in the test set), preventing the model from learning trivial mappings. We also used the driver conditions from the previous 10 days as input to predict wildfire risk on the following day.

\subsection{Dataset Analysis}

\begin{table*}[htbp]
\centering
\begin{tabular}{c|cccc|cccc}
\toprule
\multirow{2}{*}{Models} & \multicolumn{4}{c|}{w/o Position} & \multicolumn{4}{c}{w/ Position} \\ 
\cmidrule(lr){2-5}\cmidrule(lr){6-9}
{} & P & R & F1 & PR\_AUC & P & R & F1 & PR\_AUC \\ 
\midrule
SCINet	&	84.77	&	88.05	&	86.38	&	94.46	&	85.18	&	86.63	&	85.90	&	94.33	\\
TSMixer	&	85.69	&	90.39	&	87.97	&	96.24	&	86.98	&	89.29	&	88.12	&	96.10	\\
CrossLinear	&	88.04	&	87.59	&	87.81	&	96.07	&	86.31	&	89.56	&	87.90	&	95.74	\\
Crossformer	&	88.74	&	87.49	&	88.11	&	96.28	&	87.89	&	89.52	&	88.70	&	96.34	\\
FEDformer	&	82.95	&	91.18	&	86.87	&	94.93	&	85.17	&	89.82	&	87.43	&	95.23	\\
S\_Mamba	&	84.21	&	86.44	&	85.31	&	94.83	&	84.99	&	90.09	&	87.46	&	95.74	\\

\bottomrule
\end{tabular}
\caption{Comparison of model performance for next-day wildfire risk prediction using driver conditions from the previous 10 days and evaluating the effectiveness of positional embedding.}
\label{tab:results}
\end{table*}




Following the aforementioned data collection and processing procedures, our final dataset for model training, validation, and testing comprises 1,015,275 samples, including 338,425 positive samples and 676,850 negative samples. To examine the distributional characteristics and temporal dynamics of wildfire-driving factors across multiple time scales, we computed the probability density functions for wildfire and non-wildfire pixels over the preceding 365 days and analyzed their temporal trends during the last 10 days prior to ignition, as shown in Figure~\ref{fig:distribution}(a) and Figure~\ref{fig:distribution}(b), respectively. The results indicate that temperature increases, fuel accumulation, precipitation decreases, and surface soil moisture reductions are strongly associated with wildfire ignition. However, substantial distributional overlap remains between burned and unburned areas, underscoring the intrinsic stochasticity of wildfire occurrence in space and time, whereby under similar environmental conditions, fires may either ignite or remain dormant even within high-risk zones. Interestingly, regions experiencing wildfire ignition tend to exhibit slightly higher deep-layer (100–289 cm) soil water content, suggesting that wildfires often occur in ecologically moist, fuel-rich environments where deep soil moisture does not directly mitigate surface drying and ignition risk.


\section{Experiments}




\subsection{Evaluation Setups}

\subsubsection{Evaluated Models.}
We evaluated 6 time series forecasting models on the BCWildfire dataset, covering a diverse range of architectures, including CNN-based, linear-based, Transformer-based, and Mamba-based approaches. Specifically, the models include the CNN-based SCINet~\cite{liu2022scinet}, linear-based models such as CrossLinear \cite{zhou2025crosslinear}, and TSMixer~\cite{wang2024timexer}, Transformer-based models including Corssformer~\cite{zhang2023crossformer} and FEDFormer~\cite{zhou2022fedformer}, as well as the Mamba-based S\_Mamba \cite{wang2025mamba}. 


\begin{figure}[t]
    \centering
    \includegraphics[width=1\linewidth]{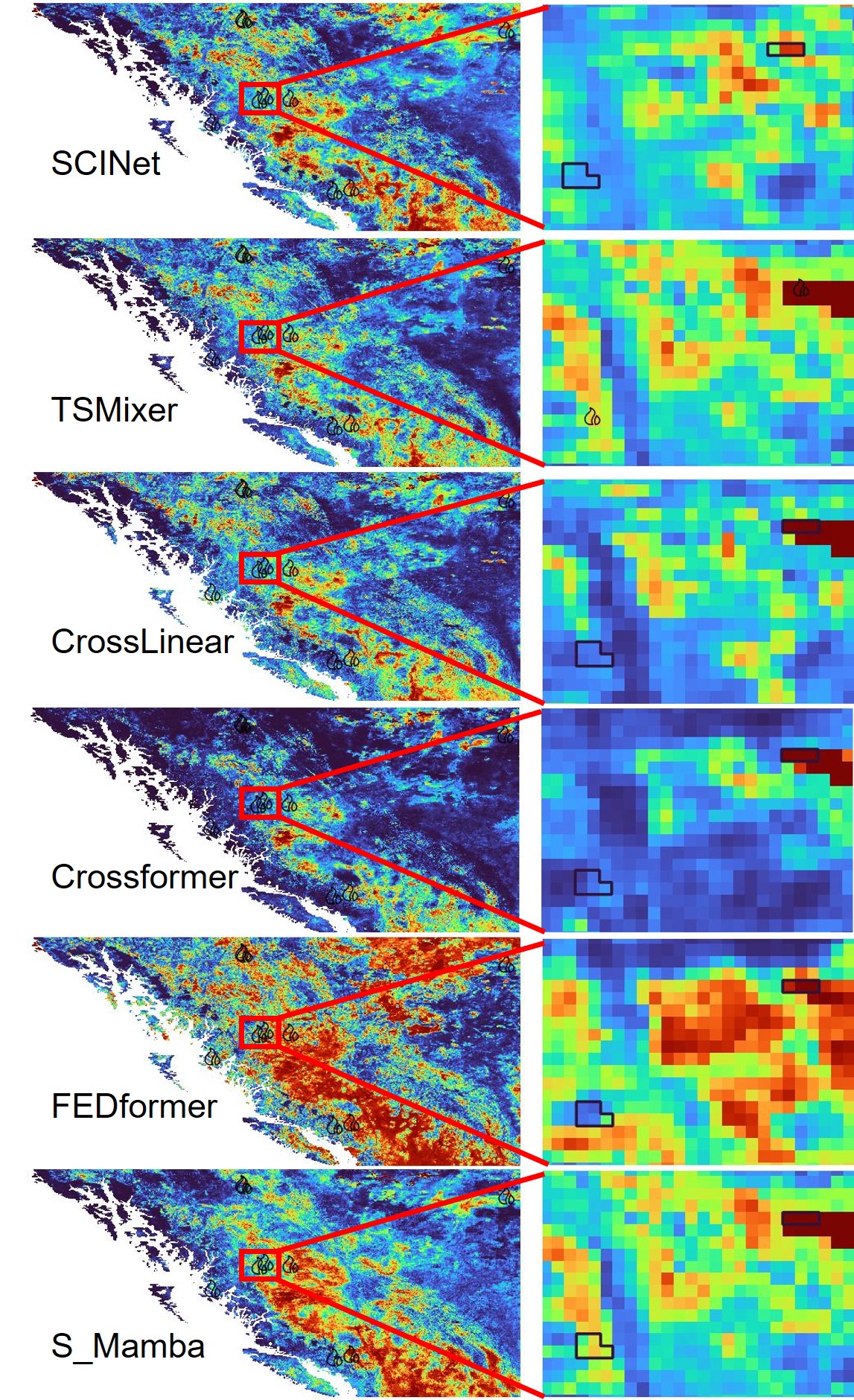}
    \caption{Visualization of predictions from different time series forecasting models on August 15, 2024.}
    \label{fig:visualization}
\end{figure}

Through extensive comparative experiments across multiple models, we aim to address several key challenges at the intersection of wildfire risk assessment and time-series forecasting. These challenges include identifying the performance limits of existing time-series models for wildfire prediction, understanding the relative importance of wildfire-driving factors within the study area, and investigating how factors such as spatial encoding influence predictive performance and model interpretability. Our findings provide clearer guidance for the effective use of datasets and the design of models in this field.

\subsubsection{Evaluation Protocols.}
We evaluate model performance using recall (R) and precision (P), F1-score (F1), and precision-recall area under curve (PR-AUC). Binary cross-entropy loss is used for model training. All models are trained using the Adam optimizer on two NVIDIA A6000 GPUs. We use a batch size of 128 and train for 50 epochs. The learning rate is scheduled to $1 \times 10^{-5}$ during training. For data partitioning, we use the years 2000-2020 as the training set, 2021-2022 as the validation set, and 2023-2024 as the test set.

\subsection{Main Results}

\subsubsection{Mechanistic Drivers of Wildfire Risk.}

To identify the short term mechanisms underlying wildfire ignition, we conducted a SHAP analysis on the Crossformer model using a 10 day input sequence to predict next day wildfire risk. The analysis indicates that ignition probability in our study region is primarily governed by recent fire activity, soil moisture dynamics, surface energy flux, and fuel conditions derived from multispectral reflectance.

The most influential feature is fire detection, whose strong negative SHAP contribution suggests that the absence of recent fire activity is a key signal for identifying potential new ignitions, while persistent fire signals correspond to already burned areas. Following this, soil moisture L3 (28–100 cm depth) and surface latent heat flux show strong positive effects, indicating that intermediate soil moisture and high surface energy exchange promote rapid drying of near surface fuels and create favorable ignition conditions. Snow cover exerts a pronounced negative influence, confirming that the presence of snow effectively suppresses ignition risk.

Spectral variables such as Band 3 (459–479 nm reflectance), Band 32 (11.77–12.27 µm emissivity), and vegetation indices including EVI and FPAR contribute positively, revealing that bright visible reflectance, high emissivity, and productive vegetation jointly reflect exposed, desiccated, and fuel rich surfaces prone to burning. Topographic variables such as hillshade, slope, and aspect rank within the top 15 predictors, suggesting that terrain controlled radiation and wind exposure further modulate ignition likelihood, although their effects remain secondary to biophysical and thermal factors.

\subsubsection*{Performance Ceilings}

Our benchmark reveals persistent performance ceilings among existing wildfire prediction architectures under the 10 day input configuration. Even the best performing models, including CrossLinear (recall 87.56\%), Crossformer (87.49\%), and FEDformer (91.19\%), plateau below 92\% recall, while precision remains within a narrow 83--88\% range. These results underscore fundamental challenges in wildfire risk prediction, such as extreme class imbalance, strong spatiotemporal heterogeneity, and complex nonlinear interactions among drivers.

Precision varies only slightly across architectures, suggesting that current representations still struggle to capture causal ignition signals and to suppress false positives. Transformer based architectures, especially Crossformer and FEDformer, achieve higher recall and stability than CNN or Mamba based models, reflecting their advantage in capturing long range temporal dependencies. Linear models such as TSMixer and CrossLinear remain competitive and efficient when temporal correlations are strong. Notably, S\_Mamba achieves balanced performance across all metrics (precision 84.21\%, recall 86.44\%, F1 85.31\%, PR-AUC 0.9483), confirming its ability to model fine scale temporal dynamics with good computational efficiency.

We further assessed the impact of incorporating spatial positional embedding, which encodes pixel level geographic information. As shown in Table~\ref{tab:results}, adding spatial context consistently improves accuracy across architectures. Transformer based models exhibit the largest gains in recall and F1 score, indicating that spatial cues enhance their ability to capture regional patterns and neighborhood dependencies. Linear and Mamba based models also benefit, with S\_Mamba showing clear improvement in recall (3.659\%) and F1 (from 2.15\%). Overall, spatial embedding provides valuable contextual information that improves model robustness and accuracy in wildfire risk prediction.

\subsubsection*{Qualitative Comparison between Different Models}

As shown on the left side of Figure~\ref{fig:visualization}, most wildfire occurrences correspond closely to areas classified as high risk. Combined with the quantitative comparison above, this result demonstrates that existing time series models can effectively predict next day wildfire risk. However, as illustrated in the zoomed in view on the right, model performance limitations remain evident, particularly in predicting new ignitions and small scale fires.

On the other hand, Figure~\ref{fig:visualization} shows that different models produce distinct spatial patterns in their predictions, particularly in false positives and high-risk regions, even though their quantitative metrics in Table~\ref{tab:results} differ only marginally. This observation suggests two key implications. First, ensemble predictions that combine different model outputs may further improve overall accuracy. Second, quantitative comparisons based solely on sampled test data may be biased and fail to fully capture the true predictive capability of each model.

\section*{Conclusion}

We present BCWildfire, a 25 year multimodal wildfire benchmark dataset covering 240 million hectares across British Columbia and adjacent northern regions. The dataset integrates 38 key wildfire driving factors, including meteorological, fuel, topographic, and human activity variables, harmonized at 1 km spatial and daily temporal resolution. Based on this resource, we establish the first unified time series forecasting benchmark for wildfire risk prediction.

Our experiments show that advanced architectures such as Crossformer, FEDformer, and S\_Mamba achieve strong predictive performance but remain constrained by the stochastic nature of ignition events and class imbalance between fire and non fire samples. SHAP analysis reveals that wildfire risk is shaped by recent fire activity, soil moisture transitions, surface energy flux, snow cover, vegetation productivity, and terrain effects, reflecting physically meaningful ignition mechanisms. Incorporating spatial positional embedding further improves accuracy across models, emphasizing the importance of spatial context in capturing regional fire dynamics.

Future work will extend the temporal horizon of modeling to better exploit the dataset’s long-term information and to examine how extended temporal drivers influence wildfire occurrence and evolution.

\section*{Acknowledgments}

This work was supported in part by the Natural Sciences and Engineering Research Council of Canada Discovery Grant (Grant No. RGPIN-2019-06744) and in part by MITACS in partnership with BMO. Zhengsen Xu acknowledges support from the China Scholarship Council.

\bibliography{aaai2026}

\end{document}